\documentclass[man,floatsintext]{apa7}

\usepackage{lipsum}

\usepackage[american]{babel}

\usepackage{csquotes}
\usepackage[style=apa,sortcites=true,sorting=nyt,backend=biber,natbib=true]{biblatex}
\usepackage{caption}
\usepackage{subcaption}
\usepackage{tabularx} 
\usepackage{multirow} 
\usepackage{comment}
\usepackage{float}

\DeclareLanguageMapping{american}{american-apa}
\title{Reducing annotator bias by belief elicitation}
\shorttitle{Reducing annotator bias by belief elicitation}

\author{Terne Sasha Thorn Jakobsen\textsuperscript{1,2}, Andreas Bjerre-Nielsen\textsuperscript{1}, Robert Böhm\textsuperscript{1,3}}
\affiliation{
\textsuperscript{1}{Copenhagen Center for Social Data Science, University of Copenhagen},
\textsuperscript{2}{Copenhagen Research Centre for Biological and Precision Psychiatry, Copenhagen University Hospital.},
\textsuperscript{3}{Department of Psychology, University of Vienna}}

\abstract{Crowdsourced annotations of data play a substantial role in the development of Artificial Intelligence (AI). It is broadly recognised that annotations of text data can contain annotator bias, where systematic disagreement in annotations can be traced back to differences in the annotators’ backgrounds. Being unaware of such annotator bias can lead to representational bias against minority group perspectives and therefore several methods have been proposed for recognising bias or preserving perspectives. These methods typically require either a substantial number of annotators or annotations per data instance. In this study, we propose a simple method for handling bias in annotations without requirements on the number of annotators or instances. Instead, we ask annotators about their beliefs of other annotators' judgements of an instance, under the hypothesis that these beliefs may provide more representative and less biased labels than judgements. The method was examined in two controlled, survey-based experiments involving Democrats and Republicans ($n=1,590$) asked to judge statements as arguments and then report beliefs about others' judgements. The results indicate that bias, defined as systematic differences between the two groups of annotators, is consistently reduced when asking for beliefs instead of judgements. Our proposed method therefore has the potential to reduce the risk of annotator bias, thereby improving the generalisability of AI systems and preventing harm to unrepresented socio-demographic groups, and we highlight the need for further studies of this potential in other tasks and downstream applications.}


\authornote{

   \addORCIDlink{Terne Sasha Thorn Jakobsen}{0000-0002-3124-1320}
   
   \addORCIDlink{Andreas Bjerre-Nielsen}{0000-0003-3057-5975}
   
   \addORCIDlink{Robert Böhm}{0000-0001-6806-0374}

  Correspondence concerning this article should be addressed to Terne Sasha Thorn Jakobsen, University of Copenhagen, E-mail: terne.jakobsen@sund.ku.dk}

\begin{document}
\maketitle

Recent advances in artificial intelligence (AI) owes its success to annotators---individuals who, given raw data such as text, images, audio or video, provide labels and other kinds of annotations describing one or more aspects of the data. Annotations are increasingly being collected via crowdsourcing platforms, where text data makes up a large part of the traffic.  
The goal of annotation has generally been to get \textit{true} labels for each instance in a dataset, under the assumption that there exists one correct interpretation. This is heavily challenged by the existence of systematic disagreement among annotators, which, in turn, challenges the ability to learn meaningful and unbiased patterns from the data and to develop unbiased AI systems, if not handled correctly. In Natural Language Processing (NLP), a research field devoted to computationally processing and understanding spoken and written language, it is broadly recognised that genuine disagreement exists across a wide range of annotation topics and tasks \citep{uma-etal-2021-semeval,plank-2022-problem}---\textit{genuine} because it does not (always) represent errors or ``noise,'' rather, it is (often) an effect of different, sometimes equally valid, interpretations. It is tempting to conclude that this challenge only pertains to ``subjective'' tasks or to non-expert, crowd-sourced annotations, but systematic disagreement is found among experts \citep{recasens-etal-2012-annotating,Aroyo2015TruthIA}, near-expert students \citep{poesio-artstein-2005-reliability}, and non-expert crowd-workers \citep{leonardelli-etal-2021-agreeing} in both subjective tasks, such as annotating toxic language \citep{sap-etal-2022-annotators}, and seemingly objective tasks, such as annotating part-of-speech (\cite{plank-etal-2014-linguistically}, see also \citealp{basile-etal-2021-need}, for more examples of disagreement in ``objective'' tasks).

What drives disagreement is not always clear, though ambiguity (within instances or instructions; \cite{jurgens-2014-analysis,plank-etal-2014-linguistically,sandri-etal-2023-dont}) and biases connected to socio-demographic attributes \citep{sap-etal-2019-risk,Kuwatly2020IdentifyingAM,liu-etal-2022-toward,fleisig-etal-2023-majority} appear to be key factors. In this study, we focus on the latter---socio-demographic annotator bias---which we define as systematic disagreement that can be operationalised as differences in annotations between groups of individuals with one or more distinctly different socio-demographic attribute(s).

Due to socio-demographic biases, attempting to construct a true label by an aggregated value, from several annotators' judgements of the same instance, is likely to erase minority-group perspectives and result in representational bias \citep{prabhakaran-etal-2021-releasing}. Several studies indicate that disagreement in annotations provides valuable information that should be used unaggregated in the development and evaluation of the systems using the data \citep{Peterson2019,Uma2020ACF,Basile2020ItsTE,basile-etal-2021-need,davani-etal-2022-dealing}. 
However, these proposals tend to require a very distinct set of annotators, who have each annotated a large amount of instances, to infer individual-specific preferences and bias, or a large amount of unique annotators and annotations per instance to infer representative label distributions for instances. Another large body of work has been focused on \textit{bias mitigation}, where the rhetoric is on removing a specific type of bias (e.g., gender bias) already present in a dataset or system, rather than preserving perspectives from disagreements (see \citealp{Hort2022BiasMF}, for a comprehensive survey of bias mitigation). 

In this study, we aim to address the important challenge of annotator bias early in the system development pipeline: during annotation. We propose a simple strategy to gain knowledge of bias in annotations \textit{without requirements on the amount of annotators or instances}. Simply, by asking annotators about their beliefs of other annotators' judgements of an instance, we can observe differences between the individual's own judgement and belief, providing a signal of bias. We hypothesise that beliefs about others' annotations may provide more representative and less biased labels than judgements, and we therefore examine the potential of our annotation method as a \textit{bias reduction} method.

\begin{figure}[h]
    \caption{}
    \begin{center}
    \includegraphics[width=0.85\textwidth]{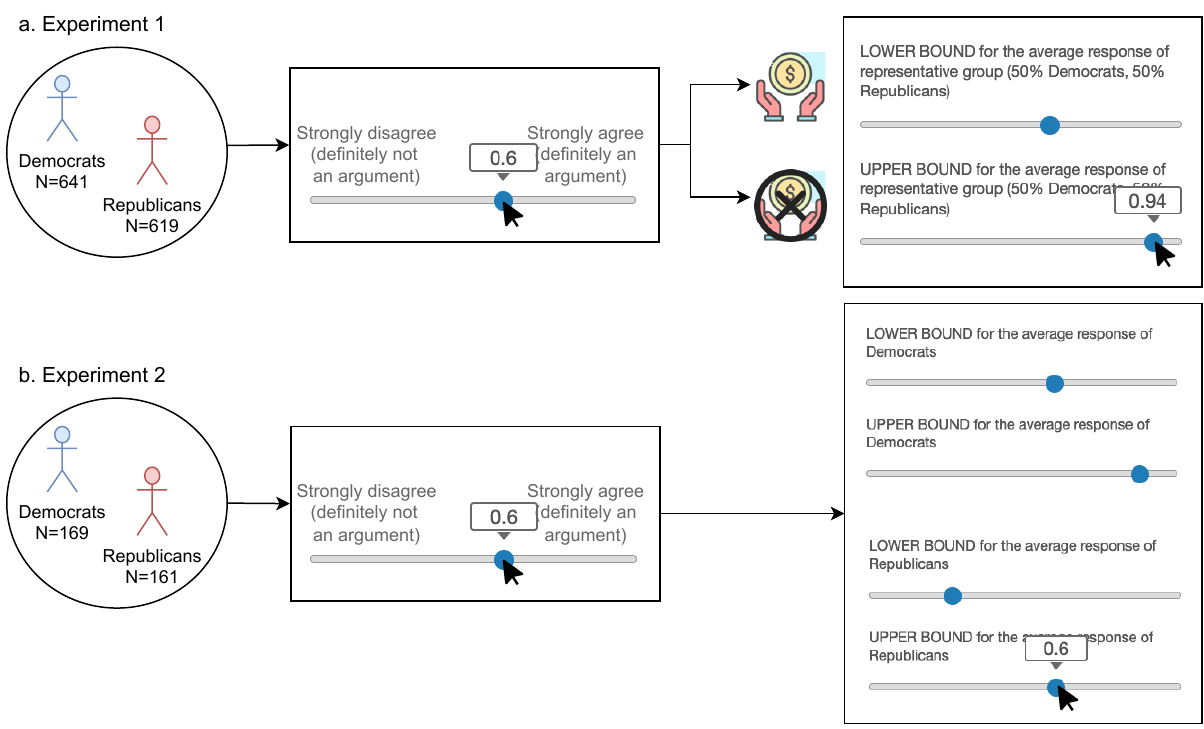}
     \end{center}
    \bigskip
    
    \small\textit{Note}. Graphical overview of the two survey-based experiments. Both experiments include two tasks (one concerning judgement and one concerning belief), where only the second differs between the experiments. \textbf{a.} In Experiment 1, the first task consisted of judging six statements (each on a separate page) wrt. whether the statements were arguments. (see statements in Table \ref{tab:arguments}). For the second task, the participants were randomly assigned to one of two conditions, one providing monetary incentives for more accurate beliefs and the other providing no monetary incentives. Belief elicitation consisted of providing an interval for the belief of what other participants (a population with 50\% Democrats and 50\% Republicans) responded to task 1. \textbf{b.} In Experiment 2, the first task was the same as in Experiment 1 but with four of the six statements to consider. All participants received the same instructions for the second task (without monetary incentives) but contrary to Experiment 1, they were asked to give two distinct intervals, one for their belief of Democrats' responses and one for their belief of Republicans' responses.
    \label{fig:process}
\end{figure}

To examine our proposed annotation method, we consider an annotation case where we know socio-demographic annotator bias exists: annotators' political affiliation. For instance, \citet{luo-etal-2020-detecting} found that annotating the stance of opinionated claims concerning global warming is biased by annotators' political affiliations (Republican, non-Republican). Similarly, \citet{thorn-jakobsen-etal-2022-sensitivity} found that recognising statements as arguments is biased by annotators' political alignment (conservative, liberal) to varying degree depending on differing annotation instructions. Here, we conducted two controlled, survey-based experiments where Democrats and Republicans are asked to judge statements as arguments, following a simple definition of what an argument is, and afterwards report beliefs about others' judgements (Figure \ref{fig:process}). We distinguish between \textit{judgement} and \textit{belief} in this way throughout the paper: a participant's \textit{judgement} is the standard annotation collected following a normal annotation procedure. A \textit{belief} is the annotation which the participant expects other participants to give. Importantly, judgement is not synonymous to opinion, rather it is the interpretation of an annotation guideline and a given sample. The experiments consist of a small set of statements and a large sample of annotators (total $N=1,590$ after exclusions) to accurately measure the effect of asking for beliefs over data instances where judgements show political bias.

\newpage

\section{Method}

The two experiments are outlined in Figure \ref{fig:process}. Experiment 1 elicited beliefs over a representative groups' responses, using a popular method for belief elicitation that usually involves providing monetary incentives. We implemented two treatment arms, one with and one without monetary incentives. Our motivation for doing this was that although monetary incentives is the norm for the method, further compensation is costly, and existing work have shown potential negative side-effects of such incentives that could harm annotation quality \citep{Gneezy2000,Benabou2006,Bowles2008}.
Motivated by the need to distinguish the annotators' beliefs over Democrats' and Republicans' responses, to better understand the compounded representative belief, Experiment 2 elicit these two beliefs separately.

\subsection{Participants}\label{participants}
Participants were recruited via Prolific (\url{www.prolific.com}). Recruitment criteria included US nationality and either Democrat or Republican political affiliation. Furthermore, the recruitment of Democrats and Republicans (to separate study links) was balanced by sex and participants who had participated in earlier studies (i.e., pilot study in the case of Experiment 1, and pilot study and Experiment 1 in the case of Experiment 2) were excluded from participating in later studies. Importantly, responses of participants who reported a different political affiliation, than they were recruited by, were excluded from analysis (80 exclusions Experiment 1; 22 exclusions Experiment 2). 

After exclusions, the experimental samples were composed of $n=641$ Democrats and $n=619$ Republicans in Experiment 1, and of $n=169$ Democrats and $n=161$ in Experiment 2. Other demographic variables included age, gender, English-speaking ability and political left-right alignment. The latter serves as another perspective on the participants' political placement, while the former serves mainly to reduce priming effects of asking for political affiliation.

\subsection{Data}\label{data}

The statements we asked participants to annotate originate from the DDO dataset (\url{https://esdurmus.github.io/ddo.html}) of \citet{durmus-cardie-2019-corpus,durmus-cardie-2018-exploring}, which is a large collection of debates and user profiles from the online debating forum \url{debate.org}. Texts were picked, for our experiments, based on the following criteria: (i) they concern controversial topics and (ii) they can be accepted as arguments given the guideline. Initially, for a pilot study, there was a third criterion: (iii) within a topic there exist two arguments expressing a Republican and a Democratic argument, where one is supporting and the other is opposing the topic. The initial selection resulted in 14 texts within 7 topics (Appendix, Table \ref{tab:DDO}). Of these 14 texts, six were identified as exhibiting biased annotations in the pilot study, and were therefore used in Experiment 1 (Appendix Table \ref{tab:sixstatements}), whereof the four statements exhibiting most biased judgements were used again in Experiment 2 (Table \ref{tab:arguments}). All texts were edited slightly to improve grammar and coherency, in an effort to make the arguments equally clear and unambiguous.

\begin{table}[h]
    \scriptsize
    \caption{
    }
    \begin{tabular}{llp{4.5in}}
    \toprule
    ID & {\sc Topic} & {\sc Argument} \\
    \midrule
        D1 & minimum wage &The minimum wage increasing will allow more people to have more money, stimulating the economy and helping citizens who are currently in poverty reach out of it, take a foothold, and stay in the middle class.\vspace{1.5mm}\\
         
         D2 & death penalty & A just society’s goal should be to protect and further the well-being of its people (and, indeed, of all people, since being just requires a lack of bias toward or against other societies). Killing people as a form of punishment does not, as a rule, serve the interest of such a society.\vspace{1.5mm}\\
         
         R1 & abortion & Abortion is morally unacceptable, and it goes against the qualities and ethics that make this country great.\vspace{1.5mm}\\
         
         R2 & gun control &More guns equals less crime. Just because crimes were committed with guns it does not mean control would work.\\
         
    \bottomrule
    \end{tabular}
    
    \bigskip
    \small\textit{Note}. We analyse the annotation of two arguments reflecting a D(emocrat) stance and two arguments reflecting a R(epublican) stance. The arguments were presented as ``statements'' for the annotators, but they were all chosen following the requirement that they, according to the authors, could in fact be defined as arguments based on the definition given to the annotators. Importantly, arguments where chosen if bias was found, on these arguments, in a pilot study.
    \label{tab:arguments}
\end{table}

\subsection{Experiment 1 procedure}

After providing informed consent with the project's aim, the possible risks, usage and storage of data, and their remuneration, participants were given a short survey on demographics. Then they were given instructions for the first task, where they were asked to specify how much they agreed that a given statement was an argument with respect to a given topic. (See instructions in the Appendix, Figure \ref{fig:Instructions1}.) Besides describing the task, the instructions also provided a description, with examples, of what an argument is and how they should judge the statements. The task consisted of six topic and statement pairs (Appendix, Table \ref{tab:sixstatements}), each presented on a separate page with the order randomised. On each page, the participants were given a 2-decimal sliding scale to use for annotation, with guiding labels on \textit{0 = Strongly disagree (definitely not an argument)}, \textit{0.25 = Somewhat disagree}, \textit{0.5 = Neither agree nor disagree}, \textit{0.75 = Somewhat agree} and \textit{1 = Strong agree (definitely an argument)}. On the top of each page there was a re-cap of the instructions along with the argument definition. Immediately after judging the sixth statement, participants were given instructions for the second task, where they had to specify their beliefs of what other participants---consisting of approximately 1000 participants, 50\% Democrats and 50\% Republicans---would on average respond in the first task. Participants were randomly assigned (50\% split) to receive one of two instructions, one simply describing how to provide intervals for their beliefs (Appendix, Figure \ref{fig:Instructions2}), and the other describing the chance to receive a bonus if their interval corresponded well to the true average, i.e. providing monetary incentives for shorter, correct intervals (Appendix, Figure \ref{fig:Instructions3}). We varied whether participants would receive monetary incentives for accurate beliefs in the belief elicitation task. Both instructions explicitly stated that all of the participants, they should form beliefs over, had received the exact same instructions as themselves for the first task. The format of the second task was the same for all participants. Again, each of the topic and statement pairs (the same as in the first task) were presented on separate pages, with instruction re-caps at the top, but for this second task the participants were given two sliding scales, one for a lower bound and one for an upper bound for the interval. The survey ended after completing the sixth statement.

\subsection{Experiment 2 procedure}
The first part of Experiment 2---consent, demographics survey \textit{and} the first task---was the same as in Experiment 1 (see above), with the exception that participants were only given four out of the six topic-statement pairs, shown in Table \ref{tab:arguments}. 

After completing the first task, participants were given instructions for the second task where they were asked to provide intervals for their beliefs of the average responses, to the first task, of other Democrat and Republican participants separately (see instructions in Appendix, Figure \ref{fig:Instructions4}). Again, they were explicitly told that the other participants had been given the same instructions for the first task as themselves, furthermore they were given an approximate number for the amount of Democrat and Republican participants. The instruction for how to provide intervals followed the wording of the instructions used in Experiment 1 that did not give monetary incentives. The four topic-statement pairs were shown on separate pages, with instructions re-cap at the top and with four sliding scales: an upper and lower bound for the belief of Democrats' and Republicans' average first task response. The granularity and labels on the sliding scales were the same for all tasks and experiments (see description in the above section). The survey ended after completing the fourth statement.

\subsection{Belief elicitation and incentivization procedure}\label{elicitation}

Getting consistent and true beliefs from individuals, through direct asking, has long been of interest of, especially, economists and psychologists as it has value for studying topics such as rationality, social preferences, peer-effects and belief-updating mechanisms \citep{Schlag2020SimpleBE}. Here, we lean on a well-studied method for eliciting beliefs known as the Most Likely Interval elicitation rule \citep{Schlag2015AMT}. An annotator was asked to provide an interval $[L,U] \subseteq [a,b]$ where they believed the average response, $x$, of a representative group would lie within. The interval $[a,b]$ was on a continuous scale, $a=0$ and $b=1$. $x$ is the average midpoint of the intervals provided by the participants. 
In Experiment 1, when utilising incentives, each participant's potential bonus was calculated after all responses were collected. If $x$ was outside a participant's interval $[L,U]$, then they did not receive a bonus. However, if $x$ was within the interval $[L,U]$, then they had a chance to receive a bonus, $S$, which was decreasing in the width, $W$, of their interval:
\[S= \left\{
    \begin{array}{ll}
      (1-\frac{W}{b-a})^g, & \mbox{if $x\in[L,U]$}.\\
      0, & \mbox{otherwise}.
    \end{array}
  \right. \]
  
How much a large $W$ should be penalised can be adjusted with the free parameter $\lambda$ in $g=\frac{1-\lambda}{\lambda}$, such that smaller $\lambda$ incentivise smaller intervals. In our study, $\lambda=0.5$. To utilise the same belief elicitation method \textit{without bonuses}, we simply explained the aim: intervals that are as narrow as possible while wide enough to feel confident that it can contain $x$.

\subsection{Statistical analyses}

We examined differences in participants' own judgements and beliefs with Wilcoxon signed-rank hypothesis tests. Differences between Democrats' and Republicans' responses were examined with one-sided Mann-Whitney U-tests. We used non-parametric tests since judgements are not normally distributed. Both types of tests were performed with the Python library Scipy. Reported hypothesis-test results are over the four arguments included in both experiments and with a mean of each participant's response for the two Democrat arguments and for the two Republican arguments. For beliefs, the mean is of the interval midpoints of the two arguments. In Experiment 2, where participants provided their beliefs over other Democrats and Republicans separately, we firstly took the midpoint of each interval and then the midpoint of the two midpoints. We supplement these analyses with Linear Mixed Effects Models (LMM) fitted on the responses to Democrat and Republican arguments separately. The models predicted either judgements or beliefs given political affiliation. Participant ID was included as an additional random effect since participants have judged, and provided beliefs for, two Democrat arguments and two Republican arguments. LMMs were also used to evaluate the effect of providing monetary incentives in Experiment 1 by fitting models to predict beliefs, again separately for Democrat and Republican arguments, based on the condition (incentives or not) and with participant ID as random effect.

We estimated the effect of varying annotator sample sizes that are closer to reality with bootstrapping. To this end, we randomly sampled 1 to 50 judgement and belief annotations of the four arguments annotated in both experiments. We did this 1000 times for each sample size and report the means. We further sampled from a Democrat and a Republican population of annotators to estimate the effect of sampling from biased participant pools. We then calculated the difference between the population means (all participants' judgements or beliefs) and the bootstrap sample means with Root Mean Squared Error (RMSE).

\section{Results}
We conducted two survey-based experiments with a total of 1,590 US-based participants (810 Democrats and 780 Republicans) to evaluate the effect of belief elicitation when annotating text data.

Monetary incentives did not have any noteworthy effects on the belief response (LMM, $p=0.07$ for D arguments and $p=0.88$ for R arguments,  see Table \ref{tab:LMMresults}), and did not result in different conclusions of hypothesis testing compared to not using incentives.

\begin{figure}[h]
     \caption{}
     \begin{subfigure}[b]{0.51\textwidth}
         \centering
         \includegraphics[width=\textwidth]{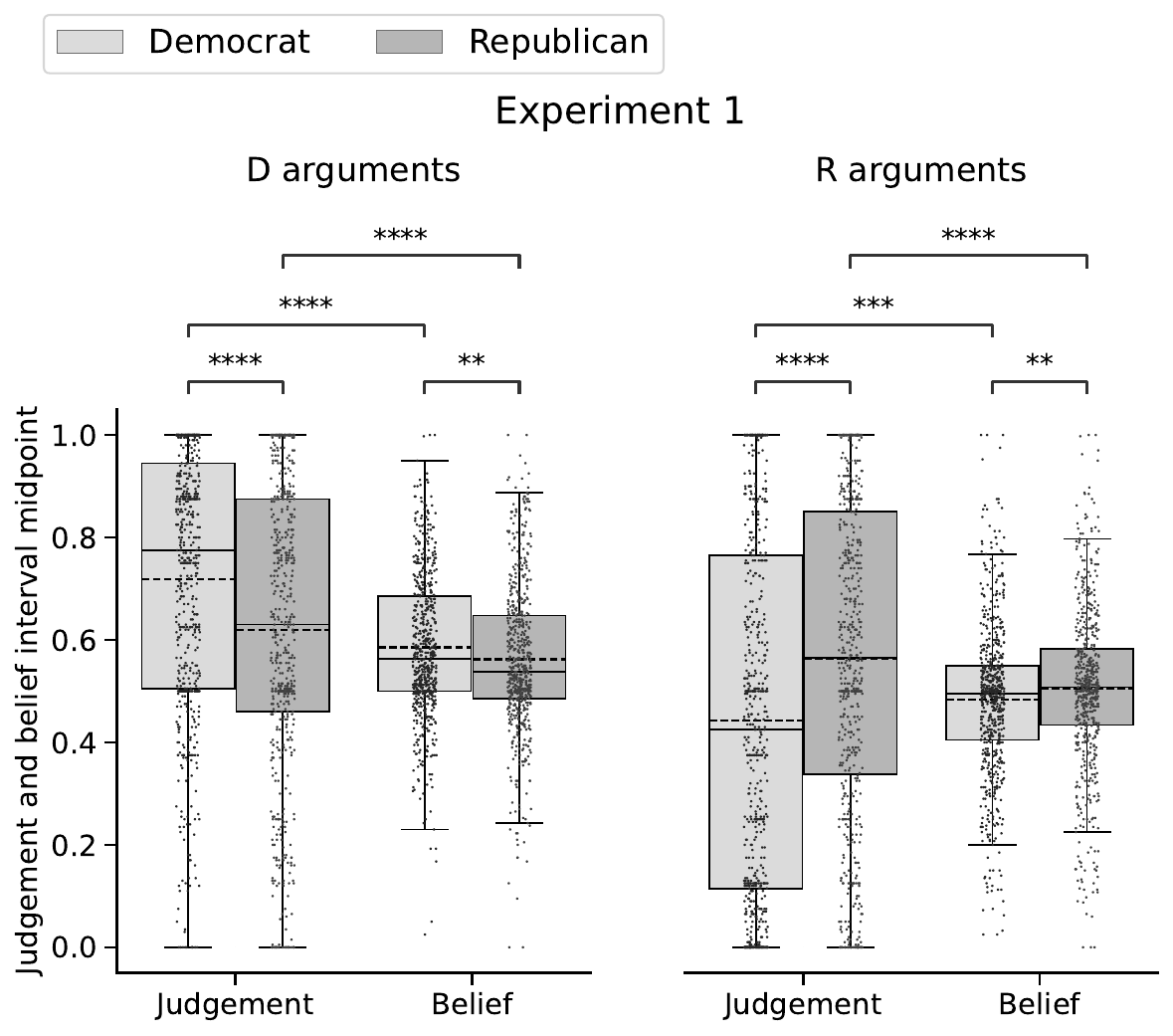}
         \label{fig:exp1}
     \end{subfigure}
     \hfill
     \begin{subfigure}[b]{0.47\textwidth}
         \centering
         \includegraphics[width=\textwidth]{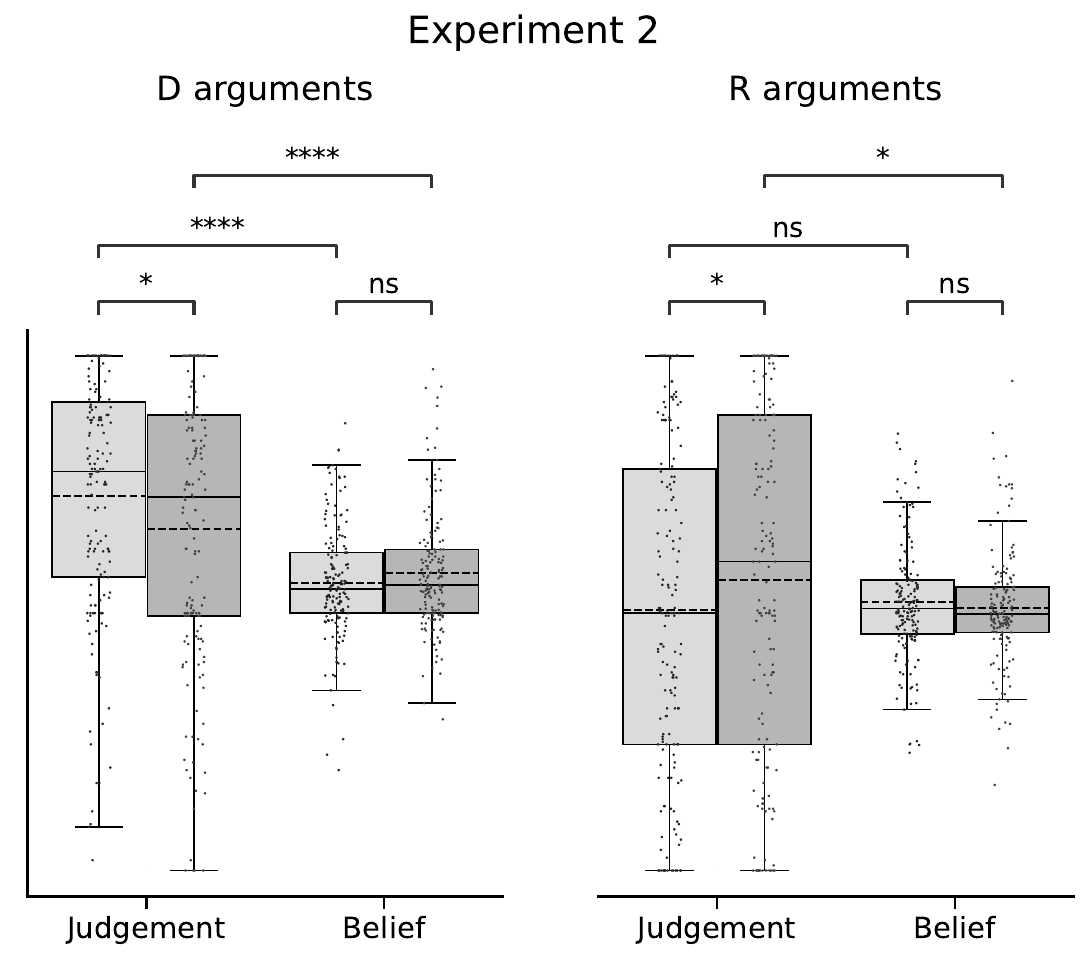}
         \label{fig:exp2}
     \end{subfigure}
    \small\textit{Note}. Boxplots showing judgements and beliefs for D(emocrat) and R(epublican) arguments in Experiment 1 (left) and 2 (right), with dashed lines indicating means. Each point shows an individual annotator's mean judgement or belief midpoint for the two D or R arguments shown in Table \ref{tab:arguments}. Stars shows significance levels of the within- and between-participants hypothesis test results shown in Table \ref{tab:res}. $^{****}p\leq$0.0001, $^{***}p\leq$0.001, $^{**}p\leq$0.01, $^{*}p\leq$0.05.
       
    \label{fig:collectedboxplots}
\end{figure}

\subsection{Judgement vs Belief}\label{awareness}
Figure \ref{fig:collectedboxplots} and Table \ref{tab:res} summarise the results of both within- and between-participants hypothesis tests for both experiments. Within-participants tests reveal that both Democrats' and Republicans' beliefs are significantly different from their own judgements, indicating an awareness of biased responses of either others or themselves. 

For Democrat annotators in Experiment 1, the directions of the adjustments from judgement to belief are as expected, with judgements of Democrat arguments being significantly higher than their beliefs, while judgements of Republican arguments are significantly lower than beliefs. This translates to being certain that arguments reflecting a Democrat stance are, in fact, arguments, while believing others think the opposite or are less certain of this fact, and vice versa for Republican arguments. Similar adjustment are seen in Experiment 2, but here, for Republican arguments, Democrat annotators' adjustments are small and insignificant.

For Republican annotators, we would expect beliefs of Democrat arguments to be an upwards adjustment from their judgements. However, in Experiment 1, Republicans' judgements of Democrats arguments are, unexpectedly, significantly higher than their beliefs, meaning they on average believe the representative group will agree less than themselves on the Democrat arguments. For Republican arguments, their judgements are, expectantly, also significantly higher than their beliefs. The same patterns are found in Experiment 2.

\begin{figure}[h]
    \caption{}
    \includegraphics[width=\textwidth]{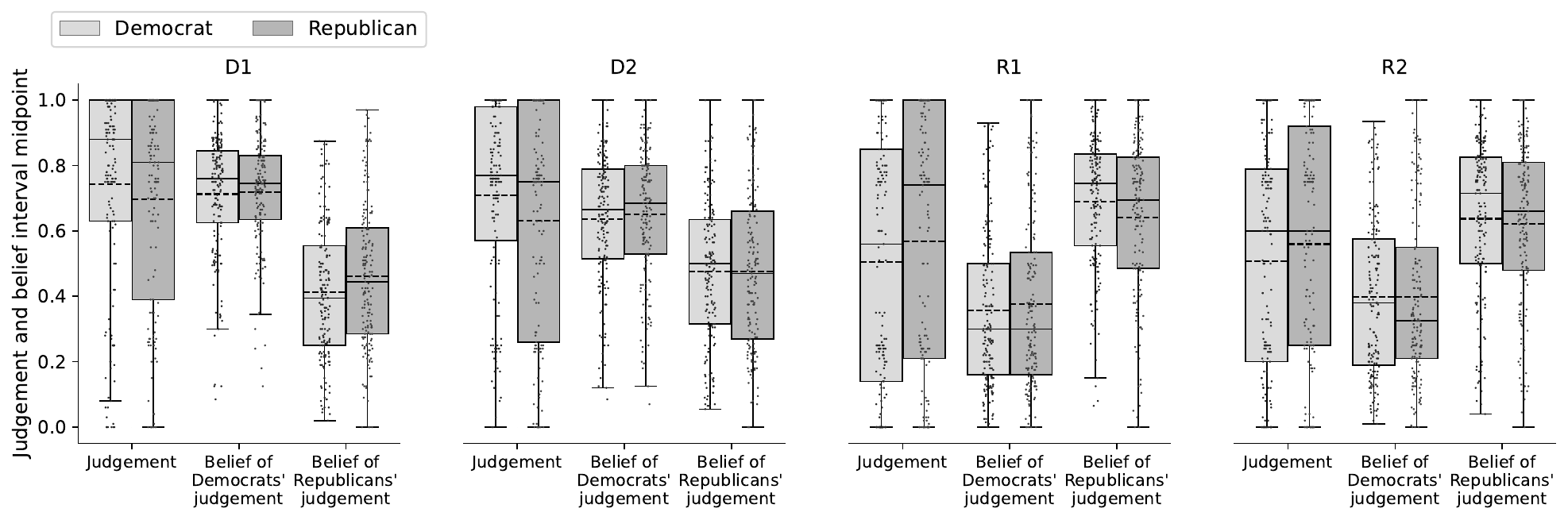}
    \small\textit{Note}. Results from Experiment 2, showing judgements and beliefs for each of the four arguments, with beliefs shown for the two separate intervals (belief of Democrats' average judgement and belief Republicans' average judgement). The results of separated beliefs show that the belief midpoints shown in Figure \ref{fig:collectedboxplots} are an effect of considering the judgement of two populations that pull the belief in two different directions.
    \label{fig:exp2beliefs}
\end{figure}

Figure \ref{fig:exp2beliefs} shows the judgements and beliefs from Experiment 2 with the annotators' \textit{separate} beliefs for other Democrat and Republican annotators' judgements. The plot illustrates how the midpoints shown in Figure \ref{fig:collectedboxplots} are the results of inferring about two populations' judgements, by clearly showing how the beliefs moves in two different directions depending on which population's judgements are being considered. This clearly indicates that the results from Experiment 1 cannot be explained by annotators always guessing around the middle of the scale, when asked for beliefs, rather than the desired effect of the separate beliefs pulling in different directions. However, for Democrat arguments, beliefs are pulled downwards and away from the elicited median annotation (i.e. they are over-adjusted) due to what seems to be wrongful or exaggerated beliefs of Republican's judgements.

\subsection{Bias reduction}
We further investigated whether beliefs are less biased than judgements by comparing the medians of Republicans' and Democrats' judgements and beliefs, separately, expecting a smaller difference between their beliefs than their (biased by construct) judgements. 

Figure \ref{fig:collectedboxplots} clearly shows that the difference between Democrats and Republicans is smaller when considering beliefs. In Experiment 1, for Democrat arguments, the difference is reduced from 0.15 (corresponding to 15 out of 100 steps on the continuous scale used for annotation) in judgements to only 0.02 in beliefs. 

Similarly for Republican arguments, the difference in medians is reduced from 0.14 to 0.01. In Experiment 2, with the smaller sample size, we see similar tendencies. For Democrat arguments, the difference in median judgement is small to begin with, yet the difference is reduced from 0.05 to 0.01. For Republican arguments, the difference is reduced from 0.10 to 0.01.

These results indicate that bias, defined as systematic differences between the two groups of annotators, is consistently reduced when asking for beliefs instead of judgements. 

\begin{figure}
    \caption{}
    \includegraphics[width=0.9\textwidth]{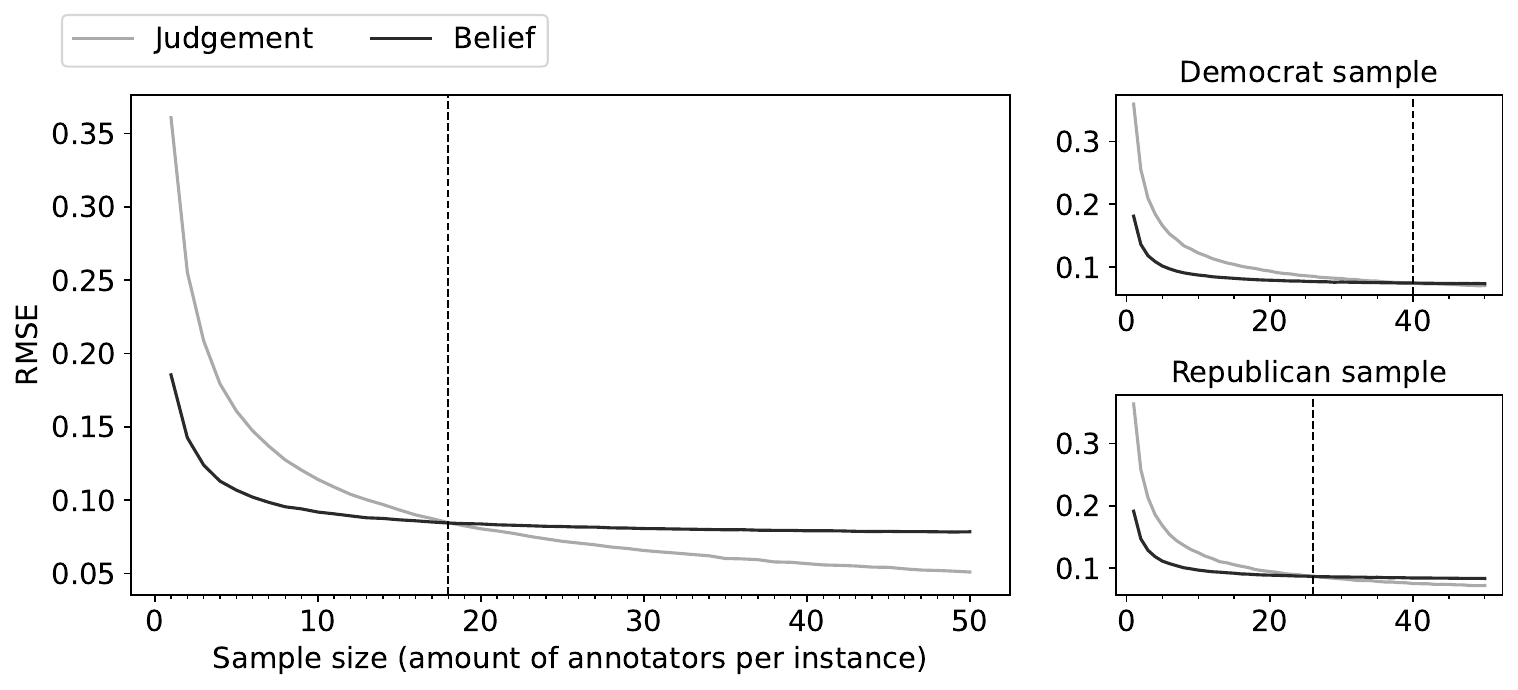}
    
    \small\textit{Note}. Result of bootstrapping judgements and beliefs with varying sample sizes and taking the RMSE between the population mean and the sample mean. The dashed line indicates the last point of which the RMSE of beliefs is smaller than the RMSE of judgements. \textbf{Left:} Belief annotations are closer to the population average when using less than 19 annotations, from 19 distinct annotators, per instance, and the smaller the sample, the larger gain in reducing bias of asking for beliefs. \textbf{Right:} The RMSE remains smaller for beliefs compared to judgements for increasingly larger sample sized when sampling from biased populations (exclusively Democrat or Republican). 
    
    \label{fig:rmsevssize}
\end{figure}

Furthermore, there is a consistently large reduction in variance, which is clearly visible in Figure \ref{fig:collectedboxplots} and confirmed with permutation tests. (Figure \ref{fig:intervals}: Supplementary Analyses). 
Hence, asking for beliefs may provide more robust annotations compared to standard practices. However, we find that, in some cases, the average belief does not accurately reflect the population average judgement, which we ultimately aim to capture. For Republican arguments, annotators over-estimate the bias of Democrats (guessing lower judgement value) and of Republicans (guessing larger judgement value). For Democrat arguments, annotators over-estimate the bias of Republicans. When the variance is large, and there is more uncertainty, as with the Republican arguments, the midpoints of beliefs provide good estimations of the population average judgement, around 0.5, signalling that there is a lot of disagreement in the instance. However, with Democrat arguments, the wrong beliefs of Republicans' judgements similarly pull the estimation towards 0.5 although it should be higher (around 0.7). The population averages of our study are, however, from a substantially larger group of annotators than a standard annotation collection for any other study (standard is 1-5 annotators per instance). We therefore ask: For which annotator sample sizes is asking for beliefs more robust than asking for judgements?  
Our bootstrapping analyses, visualised in Figure \ref{fig:rmsevssize}, shows that belief annotations are closer to the population average when collecting less than 19 annotations per instance. If sampling from unbalanced pools of annotators (either only Democrats or only Republicans), then the expected error reduction is even more robust to increasing sample sizes.

\begin{table}[h!]
\scriptsize
    \caption{}
    \begin{tabularx}{\textwidth}{l|l|XXX|XXXX}
    \toprule
        
         & & M$\pm$SD & Mdn[Quartiles] & Within $p$ &M$\pm$SD & Mdn[Quartiles] & Within $p$ & Between $p$\\
         \midrule
         &&\multicolumn{6}{c}{{\sc Experiment 1}}&\\
        & & \multicolumn{3}{c}{{\sc Democrats (n=641)}}& \multicolumn{3}{c}{{\sc Republicans (n=619)}}\\
        \midrule

         \parbox[t]{2mm}{\multirow{2}{*}{\rotatebox[origin=c]{90}{D}}} & Judgement &  0.72$\pm$0.26 & 0.78[0.51;0.95]& \multirow{2}{*}[-1pt]{4.45$\times10^{-39}$} & 0.62$\pm$0.29 & 0.63[0.46;0.88]& \multirow{2}{*}[-1pt]{2.34$\times10^{-10}$} & 5.58$\times10^{-10}$\\
         
         & Belief & 0.59$\pm$0.14& 0.56[0.50;0.69]& & 0.56$\pm$0.15 & 0.54[0.49;0.65] & & $1.42\times10^{-3}$\\
         &&& \\

         \parbox[t]{2mm}{\multirow{2}{*}{\rotatebox[origin=c]{90}{R}}} & Judgement & 0.44$\pm$0.35 & 0.43[0.12;0.77] & \multirow{2}{*}[-1pt]{6.87$\times10^{-4}$} &0.56$\pm$0.31 & 0.57[0.34;0.85]& \multirow{2}{*}[-1pt]{1.69$\times10^{-7}$} & 2.52$\times10^{-10}$\\
         
         & Belief & 0.48$\pm$0.15& 0.50[0.41;0.55] && 0.51$\pm$0.16 & 0.51[0.43;0.58] && $1.55\times10^{-3}$\\
         \midrule

         &&\multicolumn{6}{c}{{\sc Experiment 2}}&\\
         & & \multicolumn{3}{c}{{\sc Democrats (n=169)}}& \multicolumn{3}{c}{{\sc Republicans (n=161)}}\\
        \midrule

         \parbox[t]{2mm}{\multirow{2}{*}{\rotatebox[origin=c]{90}{D}}} & Judgement & 0.73$\pm$0.23 & 0.78[0.57;0.91] & \multirow{2}{*}[-1pt]{4.19$\times10^{-16}$} & 0.66$\pm$0.27 & 0.73[0.50;0.89] & \multirow{2}{*}[-1pt]{1.66$\times10^{-5}$}& 0.022\\
         & Belief & 0.56$\pm$0.11& 0.55[0.50;0.62] & & 0.58$\pm$0.12 & 0.56[0.50;0.62] & & 0.83\\
         &&& \\

         \parbox[t]{2mm}{\multirow{2}{*}{\rotatebox[origin=c]{90}{R}}} & Judgement & 0.51$\pm$0.31 & 0.50[0.26;0.78] & \multirow{2}{*}[-1pt]{0.48}& 0.56$\pm$0.34 & 0.60[0.25;0.89] & \multirow{2}{*}[-1pt]{0.026} & 0.046\\
         & Belief & 0.52$\pm$0.12 & 0.51[0.46;0.56] & & 0.51$\pm$0.11 & 0.50[0.46;0.55] & & 0.87\\

         \bottomrule
    \end{tabularx}
     \bigskip
     
    \small\textit{Note}. Statistics and hypothesis test results. \textbf{Rows}: Results shown for D(emocrat) and R(epublican) arguments, using an annotator's mean response of the two respective arguments (D1 and D2; R1 and R2), where annotators have provided both a judgement and belief. \textbf{Columns}: Separately for the two experiments and for Democrat and Republican annotators, we report Mean and Standard Deviation, Medians and quartiles [25th;75th quartile], and $p$-value of two-sided Wilcoxon Signed-Rank tests (within-participant test of difference between judgement and belief). Lastly, we report the $p$-value of one-sided Mann-Whitney U-tests (between-participant test of difference between Democrats and Republicans beliefs and between Democrats and Republicans judgements). These results are visualised in Figure \ref{fig:collectedboxplots}.
    \label{tab:res}
\end{table}

\section{Discussion}

In this paper, we have investigated a simple method for reducing bias in annotations of text data. Building on the substantial, yet sometimes overlooked, evidence of systematic disagreement in annotations of annotators with different socio-demographic backgrounds (which we call \textit{bias}), we treat differences as meaningful rather than `noisy' signals. Training models on data which does not have balanced or representative labels is harmful to models' generalisability and, if going by undetected, can be harmful to the unrepresented socio-demographic groups when such models are applied. Our study used the task of annotating arguments in text as a case where we know annotator bias exists and may result in models less able to recognise arguments that are not aligned with a specific political stance.
Common methods for handling disagreement, such as simply taking the mean of a handful annotations of the same instance, are ineffective for creating accurate and fair label distributions, and previous efforts towards improving the status-quo have required either a large pool of annotators or instances per annotator, which might not be feasible in all projects. 

We show that by changing the way we ask for annotations, we can get labels that are more representative of diverse backgrounds. Specifically, annotators' beliefs of other peoples' annotations provide labels closer to the average judgement of a large, representative group of annotators. This holds when using less than 19 annotators per instance, or up to 40 if recruiting an unbalanced, or unrepresentative, group of annotators for the task of annotating arguments in text. 

Asking for both a judgement and belief annotation, and utilising signals of uncertainty given by both (and in the difference between them), may have broader implications for improving model generalisability. \citet{Peterson2019} and \citet{Uma2020ACF} show that utilising human uncertainty in annotations can improve the generalisability of systems, and \citet{basile-etal-2021-need} argue that it can make the evaluations more stable. While previous studies infer uncertainty from the annotations of several individuals for each instance, our annotations provide cheaper, individual-level uncertainty measures. 

A final and crucial aspect of belief annotations is the significant reduction in variance. While it is essential to keep diversity in perspectives to develop unbiased systems, large variance in annotations undoubtedly makes it more difficult for ML systems to learn meaningful patterns. Usually, low-bias in ML comes in a trade-off with high-variance. However, we found that asking for beliefs produce annotations that are at the same time more representative of the specified population and has significantly less variance. Belief annotations therefore have the potential for making it easier for systems to recognise and learn from instances that are dividing, and have a high degree of uncertainty, by producing low-bias, low-variance targets.

\subsection{Limitations and Future Directions}
Possible limitations of asking for beliefs are the need for being able to make good assumptions about which socio-demographic characteristics are important, as well as describing these characteristics for the annotators. In our experiments, we define the characteristics of the population(s) of which we ask annotators to estimate judgements (50\% Democrats and 50\% Republicans, or Democrats and Republicans separately). Our results suggest that the definitions are essential for the beliefs. The extend to which very specific versus broad (for instance, defining the population as representative without mentioning party affiliation) definitions influence the annotations is still left to be explored. Relatedly, our proposed method requires knowledge about the annotators' socio-demographic characteristics that may be responsible for biased annotations in the first place. If such biases have been documented and are known (i.e., ``known unknowns'', \citep{bail_2023}), our method may help to reduce bias preemptively, but it is not suitable to reduce annotation bias due to unknown annotator characteristics (i.e., ``unknown unknowns''). 
Socio-demographic characteristics only partly explain annotator biases and the extend to which such characteristics affect annotations depends on the task. \citet{fleisig-etal-2024-perspectivist} highlight that factors such as media usage and opinions may be more influential than demographics, for some tasks. (In this study, we treat political affiliation, which is closely connected to opinion, as a socio-demographic factor, while \citet{fleisig-etal-2024-perspectivist} treats opinions as seperate from demographics.)

Measuring bias, the effect of belief elicitation, and having a reliable population average, has required the recruitment of a large amount of annotators. The cost of this has been a modest set of instances which our annotation method has been tested on. Furthermore, the belief elicitation method was only tested on instances where bias was found, since we expect believe elicitation to have little or no influence on instances that are `easy' or un-divisive. 
This study serves as a proof-of-concept, showing that annotator bias, and variance, can be reduced by belief elicitation. Future directions lies in applying the method to larger, diverse datasets and training models on the annotations.

\subsection{Conclusions}

We found that eliciting beliefs in annotations has the potential to reduce the risk of annotator bias. The method is low-cost and applicable to many annotation tasks were annotation bias is expected, and to test hypotheses of the occurrence of bias. Unlike most methods for mitigating biases, belief elicitation tackles bias ex ante, rather than ex post, and does not require a large amount of annotators or instances to be applicable. We encourage further studies to investigate the method with other tasks and down-stream applications.

\section{Ethics declarations}

\paragraph{Broader impact} We hope this study will promote further research on understanding annotator behaviour and ways to reduce demographic-dependent biases \textit{before} modelling, by means of better understanding the annotators we recruit. Our study contributes to research on improving data quality by arguing for a more nuanced look on what quality means -- quality is not simply achievable by removing `outliers' or by taken the average of a handful judgements. However, quality may be improved by changing the way we ask for annotations and having more ways of inferring about annotators' beliefs and uncertainty of instances. 

\paragraph{Personal and sensitive data}
This study deals with personal and sensitive data, i.e. political affiliation, age, gender and education level. Responses are anonymous and cannot be used to identify any individual. The study has an IRB approval and both experiments were pre-registered on OSF.

\paragraph{Consent to participate}
Participants were informed of the study's objective and consented to the sharing of their anonymous responses, for research purposes. In the consent form, they were warned the texts were related to controversial topics and could include offensive statements or language.

\paragraph{Consent for publication}
Participants consented to having their anonymised data published and used for research purposes.

\paragraph{Potential risks}
While we do not anticipate any risks from participation in the study, we do note a recent awareness of poor working conditions among crowd-workers \citep{AIethics}. Since our participants are from the US and are recruited through Prolific, who have fair minimum payment rules, we do not anticipate such poor working conditions, among our participants, as are sometimes the reality for AI data annotators.

\paragraph{Costs and remuneration}
The cost of the entire study (Experiment 1 and 2) was 1987.8\textsterling\space ($\approx$ 2222\$), whereof 129\textsterling\space constitutes bonuses given in Experiment 1. Participants were paid 6.6\textsterling/h ($\approx$7.4\$/h) on average across both experiments (excluding bonuses).

\paragraph{Data and code availability}

Data and code is available on the Open Science Framework\footnote{\url{https://osf.io/d29vw/?view_only=0f550067a004429d8b6ef9ee67697aae}} and \url{https://github.com/terne/belief-elicitation}.

\paragraph{Pre-registration}
The study was pre-registered on the Open Science Framework.\footnote{Experiment 1: \url{https://osf.io/w8m45/?view_only=0b7a4df0e0f2426399a51eb4ffca8d7c}}\footnote{Experiment 2: \url{https://osf.io/4x9dz/?view_only=8f9541452538467fb31380a30a3c07ca}} See Appendix E for more details.

\paragraph{Funding}
There was no external funding for this research.

\paragraph{Competing interests}
The authors have no relevant financial or non-financial interests to disclose.

\section{Acknowledgements}

We first and foremost thank the participants, without whom the study would not have been possible. The first author extends their heartfelt thanks to  Anders Søgaard and David Dreyer Lassen for their excellent supervision during the PhD study, in which this paper took shape.

\printbibliography

\appendix

\section{Stimuli}
\label{app:stimuli}

\begin{table}[h]
    \caption{}
    \small
    \begin{tabular}{llp{4in}}
    \toprule
    ID & {\sc Topic} & {\sc Argument} \\
    \midrule
        D1 & minimum wage &The minimum wage increasing will allow more people to have more money, stimulating the economy and helping citizens who are currently in poverty reach out of it, take a foothold, and stay in the middle class.\vspace{1.5mm}\\
         
         D2 & death penalty & A just society’s goal should be to protect and further the wellbeing of its people (and, indeed, of all people, since being just requires a lack of bias toward or against other societies). Killing people as a form of punishment does not, as a rule, serve the interest of such a society.\vspace{1.5mm}\\

         D3 & global warming & CO2 is the largest contribution to global warming, at 72\% of greenhouse gas emissions, and greenhouse gas emissions are a major contributor to global warming.\vspace{1.5mm}\\
         
         R1 & abortion & Abortion is morally unacceptable, and it goes against the qualities and ethics that make this country great.\vspace{1.5mm}\\
         
         R2 & gun control &More guns equals less crime. Just because crimes were committed with guns it does not mean control would work.\vspace{1.5mm}\\

         R3 & global warming & The media coverage on pollution affecting global warming on a grand scale is a scam lead by liberals such as Al Gore, Michael Moore and the liberal media.\vspace{1.5mm}\\

    \bottomrule
    \end{tabular}
     \bigskip
     
    \small\textit{Note}. The six statements used in experiment 1.
    \label{tab:sixstatements}
\end{table}

\section{Annotation Guidelines}
We present four different annotation guidelines/instructions given to the annotators. One for the first task, which was the same for both experiments, and three for the second task (for Experiment 1 with incentives and without incentives and for Experiment 2).

\begin{figure}[h]
    \caption{}
     \begin{subfigure}[b]{\textwidth}
         \includegraphics[width=0.8\textwidth]{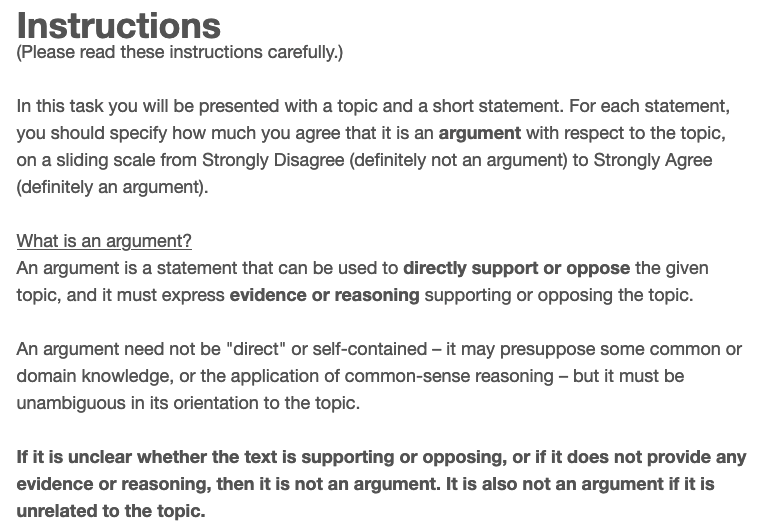}
     \end{subfigure}
     \begin{subfigure}[b]{0.78\textwidth}
         \includegraphics[width=\textwidth]{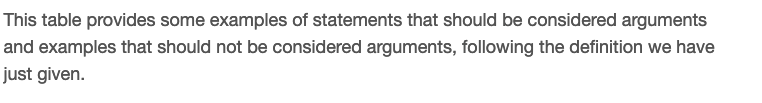}
     \end{subfigure}
      \begin{subfigure}[b]{0.7\textwidth}
         \includegraphics[width=\textwidth]{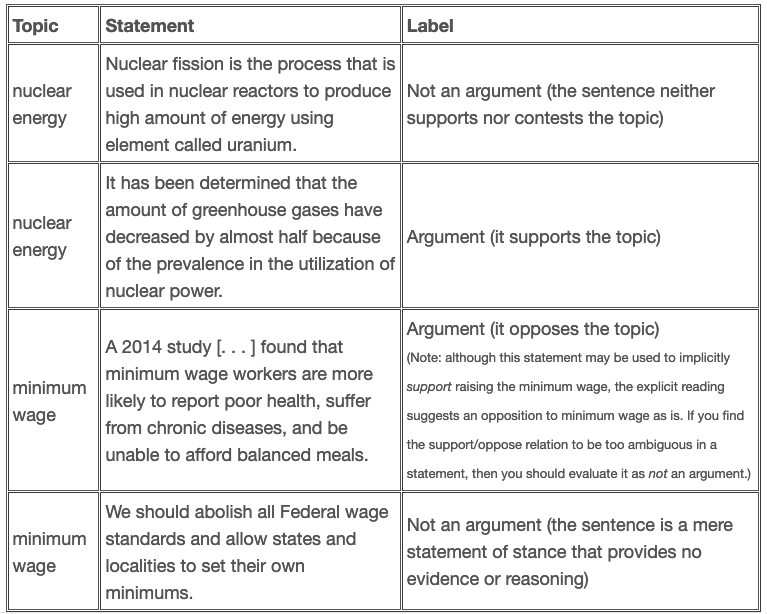}
     \end{subfigure}
     
    
    \small\textit{Note}. Task 1 instructions (for Experiment 1 and 2).
    \label{fig:Instructions1}
\end{figure}

\begin{figure}[h]
     \caption{}
     \begin{subfigure}[b]{\textwidth}
         \includegraphics[width=0.8\textwidth]{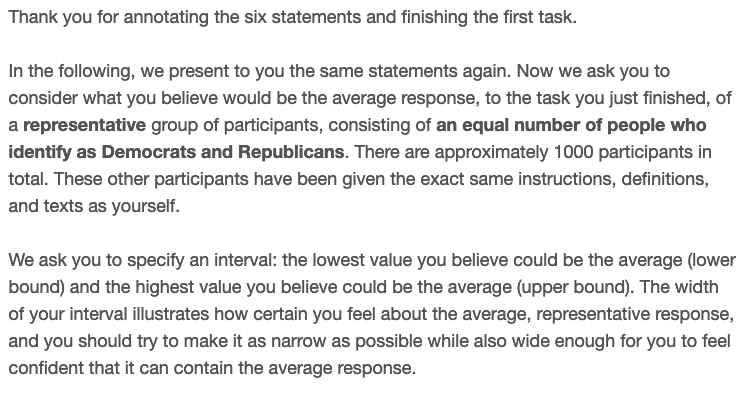}
     \end{subfigure}
      \begin{subfigure}[b]{0.8\textwidth}
         \includegraphics[width=\textwidth]{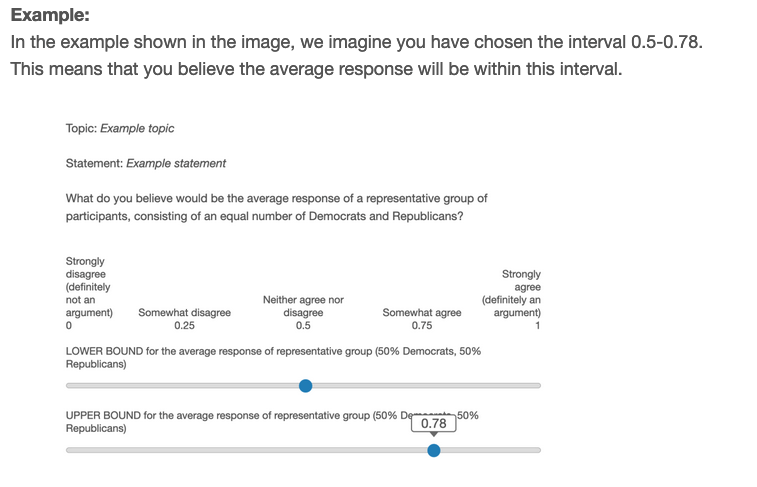}
     \end{subfigure}
    
    \small\textit{Note}. Task 2 instructions for Experiment 1 without incentives.
    \label{fig:Instructions2}
\end{figure}

\begin{figure}[h]
     \caption{}
     \begin{subfigure}[b]{\textwidth}
         \includegraphics[width=0.8\textwidth]{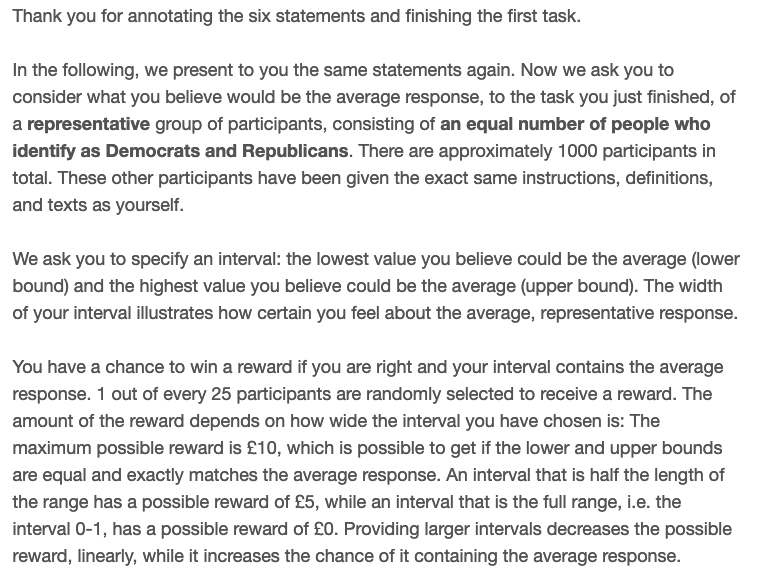}
     \end{subfigure}
      \begin{subfigure}[b]{0.8\textwidth}
         \includegraphics[width=\textwidth]{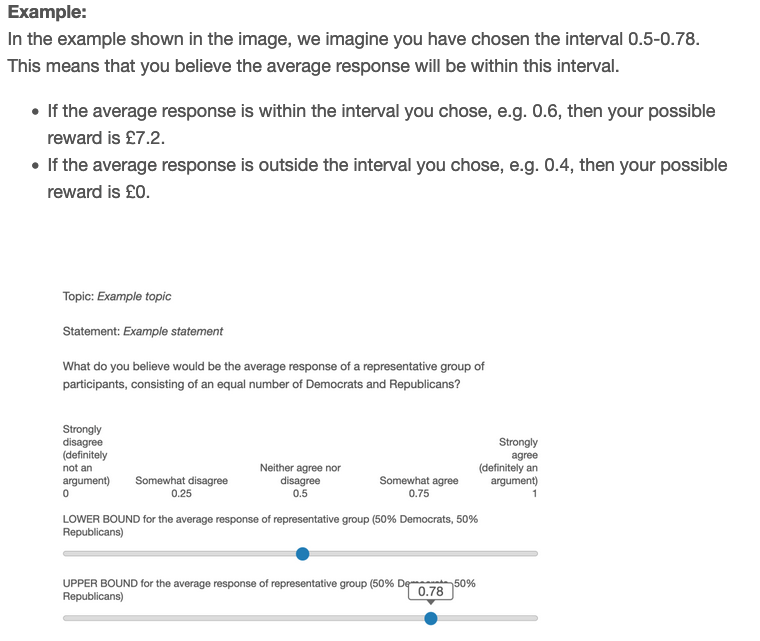}
     \end{subfigure}
    
    \small\textit{Note}. Task 2 instructions for Experiment 1 with incentives.
    \label{fig:Instructions3}
\end{figure}

\begin{figure}[h]
     \caption{}
     \begin{subfigure}[b]{\textwidth}
         \includegraphics[width=0.7\textwidth]{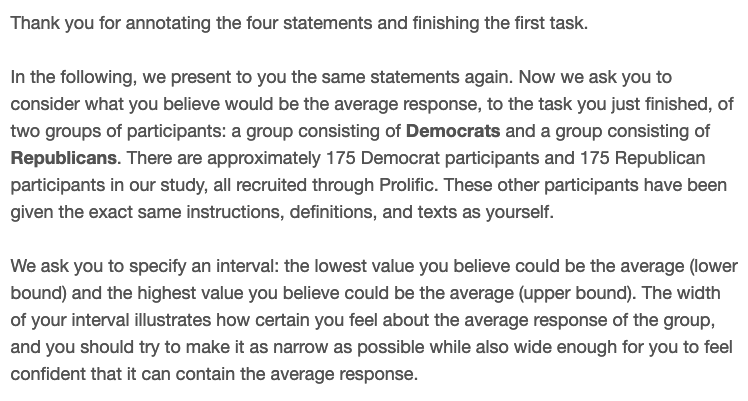}
     \end{subfigure}
     \begin{subfigure}[b]{0.7\textwidth}
         \includegraphics[width=\textwidth]{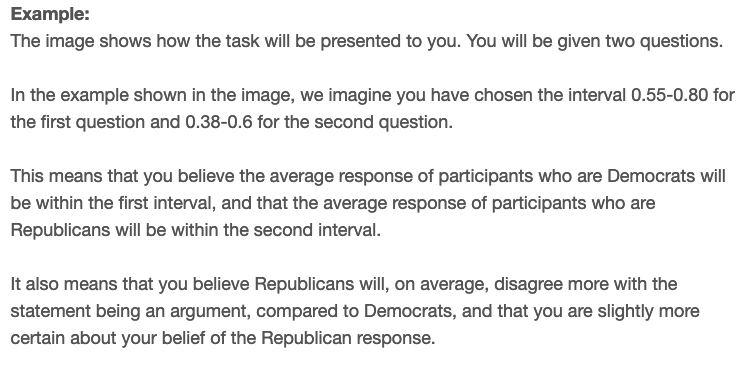}
     \end{subfigure}
      \begin{subfigure}[b]{0.7\textwidth}
         \includegraphics[width=\textwidth]{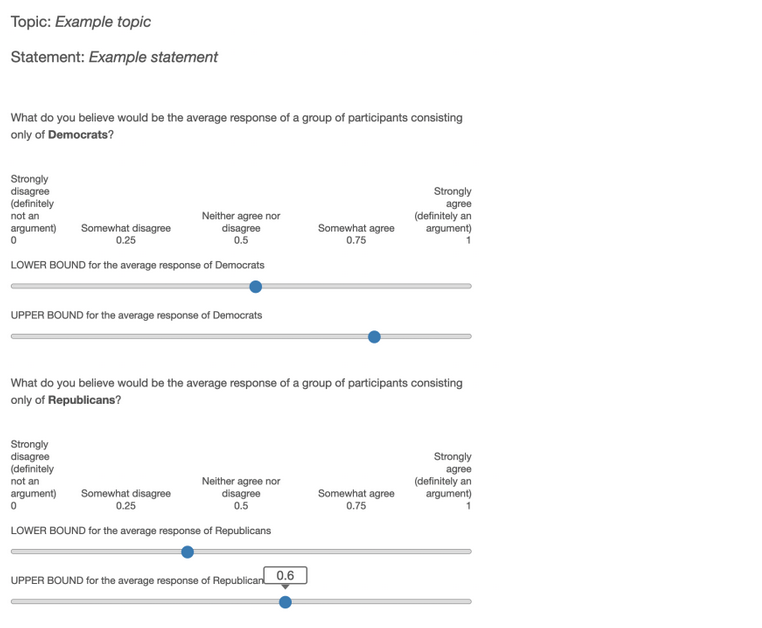}
     \end{subfigure}
    
    \small\textit{Note}. Task 2 instructions for Experiment 2.
    \label{fig:Instructions4}
\end{figure}

\section{Pilot}
\label{app:pilot}

20 Democrats and 20 Republicans were recruited (via Prolific) to participate in a pilot study where they were given 14 statements/arguments to annotate (following the Task 1 instructions, see Appendix, Figure \ref{fig:Instructions1}). Arguments were picked from the DDO dataset\footnote{ \url{https://esdurmus.github.io/ddo.html}} of \citet{durmus-cardie-2019-corpus} following three criteria: 1) they concern controversial topics, 2) they can be accepted as arguments given the guideline above, 3) within a topic there exist two arguments expressing a republican and a democratic argument, where one is supporting and the other is opposing the topic. See Table \ref{tab:DDO}. Based on the pilot results, 6 statements were picked for the following experiment. The pilot results were further used for calculating power and sample size for experiment 1.

\begin{table*}[h!]
\caption{}
\scriptsize
    \begin{tabular}{cp{2.7in}c|cc}
    \toprule
    Writer's affiliation & Statement & Topic & H & $p$ \\
    \midrule
    Republican Party & Abortion is morally unacceptable, and it goes against the qualities and ethics that make this country great. &  abortion & D$<$R  & \textbf{0.0001}\\
    \midrule
    Democratic Party &  I feel that a woman can do whatever she wants with her body. &  abortion & D$>$R  & 0.8778 \\
    \midrule
    Republican Party & More guns equals less crime. Just because crimes where committed with guns it does not mean control would work. &  gun control & D$<$R  & \textbf{0.0098}\\
    \midrule
    Democratic Party & Background checks are a way that the government keeps guns out of known criminals' hands and people with mental health issues, without infringing upon the rights of normal citizens. & gun control & D$>$R & 0.2237\\
    \midrule
    Republican Party & Minimum wage laws are not only unnecessary but counterproductive because competition for workers keeps wages up and wage controls discourage hiring. &  minimum wage & D$<$R  &  0.9819\\
    \midrule
    Democratic Party & The minimum wage increasing will allow more people to have more money, stimulating the economy and helping citizens who are currently in poverty reach out of it, take a foothold, and stay in the middle class. &  minimum wage & D$>$R & \textbf{0.0066}\\
    \midrule
    Republican Party & Gay Marriage cannot be legislated as legal because the laws associated to governing it are outside the declaration of Marriage. & same-sex marriage & D$<$R  & 0.3525\\
    \midrule
    Democratic Party & I believe that gay marriage is acceptable, and disallowing it is discriminatory. & same-sex marriage & D$>$R  & 0.4792\\
    
    \bottomrule
    \end{tabular}
    
    \bigskip
    \small\textit{Note}. \textit{Continues on next page.}
    \label{tab:DDO}
\end{table*}


\begin{table*}[h!]
\scriptsize
    \begin{tabular}{cp{2.7in}c|cc}
    \toprule
    Writer's affiliation & Statement & Topic & H & $p$ \\
    \midrule
    Republican Party & We do not need feminism in this country anymore for it has done some good things but has evolved to the point that all it does now is degrading males and trying to get more rights for women while men get less rights. &  feminism is necessary & D$<$R  & 0.1265\\
    \midrule
    Democratic Party &  Women's bodies are treated as a commodity here in the west, many ads show women in scantily clad clothing often in suggestive poses. Sexual objectification dehumanizes women in both men and women's eyes a woman becomes just an object and not a person so a man feels free to treat her as one. &  feminism is necessary & D$>$R & 0.0969\\
    \midrule
    Republican Party & I think that the death penalty makes much more sense than does solitary confinement for life because the taxpayers of the country that the criminal broke the laws of have to suffer from paying for food, medicine, and security for said criminal. &  death penalty & D$<$R  & 0.4969\\
    \midrule
    Democratic Party &  A just society's goal should be to protect and further the wellbeing of its people (and, indeed, of all people, since being just requires a lack of bias toward or against other societies). Killing people as a form of punishment does not, as a rule, serve the interest of such a society. & death penalty & D$>$R & \textbf{0.0138}\\
    \midrule
    Republican Party & The media coverage on pollution affecting global warming on a grand scale is a scam lead by liberals such as Al Gore, Michael Moore and the liberal media. & global warming & D$<$R  & \textbf{0.0234}\\
    \midrule
    Democratic Party & CO2 is the largest contribution to global warming, at 72\% of greenhouse gas emissions, and greenhouse gas emissions are a major contributor to global warming. & global warming & D$>$R & \textbf{0.0082}\\
    \bottomrule
    \end{tabular}
    
    \bigskip
    \small\textit{Note}. \textit{Continued.} Curated dataset of seven statement pairs used in the pilot study. On the right we show the results of one-sided Mann-Whitney U-tests comparing the judgements of Democrats and Republicans of the pilot study. $p$-values in bold are significant and highlights the sentences chosen for further experiments.
\end{table*}


\section{Supplementary Analyses}

\begin{table}[]
\scriptsize
    \caption{}
    \begin{tabularx}{0.68\textwidth}{llcccX}
    \toprule
         Model& &  Coef. & Std.Err. &z-score &p-val. \\
         \midrule
         \multicolumn{6}{c}{{\sc Experiment 1}}\\
         \midrule
         \multicolumn{2}{l}{Republican$\rightarrow$}&&  &  &    \\
         &Judgement of D arguments& -0.01  & 0.02 & -6.48 & 8.92$\times10^{-11}$ \\
         
         &Belief of D arguments & -0.02 & 0.01 & -2.85 & 0.004 \\
         
         &Judgement of R arguments& 0.12 & 0.02 & 6.36 & 1.97$\times10^{-10}$ \\
         
         &Belief of R arguments & 0.02 & 0.01 & 2.45 & 0.01 \\
         
         \multicolumn{2}{l}{Incentives$\rightarrow$}&&  &  &    \\
         & Belief of D arguments & 0.02 & 0.01 & 1.84 & 0.07\\
         & Belief of R arguments &  -0.001& 0.01 & -0.16 & 0.88\\
         \midrule
         \multicolumn{6}{c}{{\sc Experiment 2}}\\
         \midrule
         \multicolumn{2}{l}{Republican$\rightarrow$}&&  &  &    \\
         &Judgement of D arguments& -0.06 & 0.03& -2.26& 0.02 \\
         
         &Belief of D arguments & 0.02 & 0.01& 1.44 & 0.15  \\
         
         &Judgement of R arguments& 0.06 & 0.04& 1.60& 0.11 \\
         
         &Belief of R arguments & -0.01 & 0.01& -0.90& 0.37  \\

    \bottomrule
    \end{tabularx}
     \bigskip
     
    \small\textit{Note}. Results of Linear Mixed Effects Models (LMM) for the effect of political affiliation on the judgements and beliefs of D(emocrat) and R(epublican) arguments. Participant IDs are treated as random effects.
    
    \label{tab:LMMresults}
\end{table}

\begin{figure}
    \caption{}
    \includegraphics[width=0.9\textwidth]{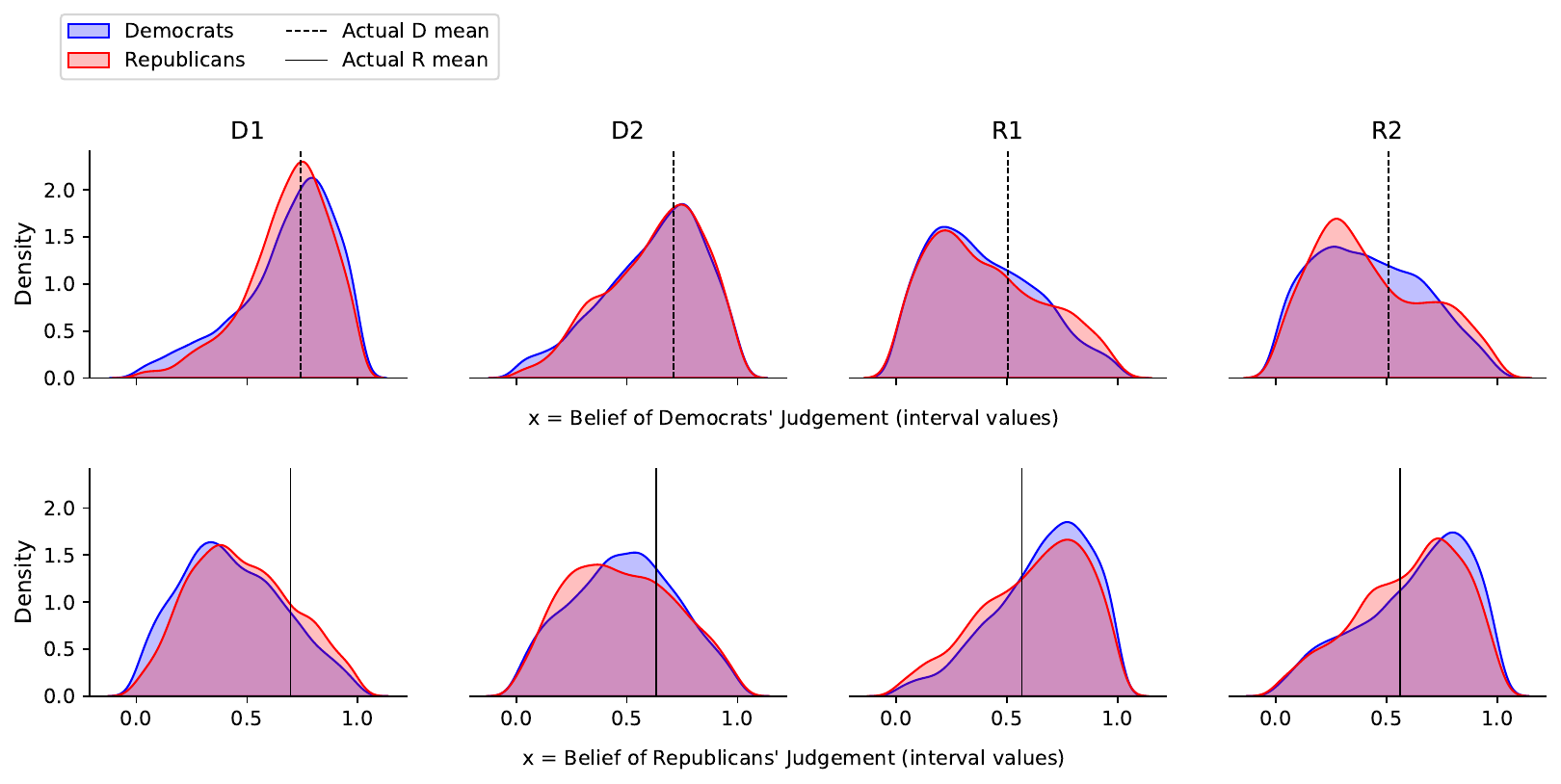}
    
    \bigskip
    \small\textit{Note}. Distribution of belief intervals from Experiment 2, constructed by concatenating annotators' lists of interval values ranging from (and including) their individual lower bound to their individual upper bound. The first row shows the intervals for the belief of Democrats' judgements, and the second row shows the intervals for the belief of Republicans' judgements.
    \label{fig:intervals}
\end{figure}

\section{Pre-registration}

Both experiments were pregistered on the Open Science Framework (OSF)\footnote{Experiment 1: \url{https://osf.io/w8m45/?view_only=0b7a4df0e0f2426399a51eb4ffca8d7c}}\footnote{Experiment 2: \url{https://osf.io/4x9dz/?view_only=8f9541452538467fb31380a30a3c07ca}}. Both pre-registrations included two hypothesis, the first of which was identical in the two experiments:

\begin{itemize}
    \item Experiment 1 and 2, Hypothesis 1: When asked about their own opinions regarding annotations of text, Democrats and Republicans differ in their annotations (annotator-bias hypothesis).
    \item Experiment 1, Hypothesis 2: When asked about their belief of how a representative population of annotators would annotate the text, responses of Democrats and Republicans move toward the population average and, thus, annotator bias decreases (belief-elicitation hypothesis). 
    \item Experiment 2, Hypothesis 2: When asked about their belief of how a Democrat and a Republican population of annotators would annotate the text, the midpoint of the two beliefs, of Democrats and Republicans, move toward the population average and, thus, annotator bias decreases (belief-elicitation hypothesis).
\end{itemize}

Preregistered \textit{confirmatory analyses} for hypothesis 1 included between-participant comparisons of judgements with non-parametric Mann-Whitney U-tests, expecting a significant difference between Democrats' and Republicans' judgements (confirmed) and within-between participants comparisons with parametric Linear Mixed Effects Models expecting a significant effect of political affiliation on judgement (confirmed in Experiment 1 and for half of the arguments in Experiment 2). For hypothesis 2, pre-registered confirmatory analyses included non-parametric within-participant comparisons of judgements and beliefs, expecting a significant difference between participants' own judgements and beliefs, in the sense that the belief is closer to the population average (confirmed difference, but not always closer to the population average), and within-between participants comparisons with parametric Linear Mixed Effects Models on the effect of political affiliation on beliefs, expecting it to differ from the effect on judgements (partly confirmed, the effect of political affiliation on beliefs is still significant in Experiment 1).

Preregistered \textit{secondary (exploratory) analyses} included between-participants comparison of beliefs, with Mann-Whitney U-tests, to see if the bias had only been reduced or completely eliminated, in which case there should be no significant difference between Democrats' and Republicans' beliefs (bias seems reduced but not eliminated), and analysis of variance reduction from judgements to beliefs (confirmed reduction in variance) and analyses of the difference between reduced annotator bias in the belief elicitation condition with incentives vs. without incentives in Experiment 1 (finding no significant effect of incentives).

\end{document}